# Enhancing Rock Image Segmentation in Digital Rock Physics: A Fusion of Generative AI and State-of-the-Art Neural Networks

**Authors:**

Zhaoyang Ma[a*]; Xupeng He[b]; Hyung Kwak[b]; Jun Gao[b]; Shuyu Sun[a]; Bicheng Yan[a]

[a]Earth Science and Engineering, Physical Science and Engineering Division, King Abdullah University of Science and Technology (KAUST)

[b]Saudi Aramco EXPEC Advanced Research Center, Reservoir Engineering Technology Division, Pore Scale Physics Focus Area, Bldg. 2291, GA-168

* Corresponding Author: Tel.: +966 563677285; E-mail address: zhaoyang.ma@kaust.edu.sa

**Abstract:**

In digital rock physics, analysing microstructures from CT and SEM scans is crucial for estimating properties like porosity and pore connectivity. Traditional segmentation methods like thresholding and CNNs often fall short in accurately detailing rock microstructures and are prone to noise. U-Net improved segmentation accuracy but required many expert-annotated samples, a laborious and error-prone process due to complex pore shapes. Our study employed an advanced generative AI model, the diffusion model, to overcome these limitations. This model generated a vast dataset of CT/SEM and binary segmentation pairs from a small initial dataset. We assessed the efficacy of three neural networks: U-Net, Attention-U-net, and TransUNet, for segmenting these enhanced images. The diffusion model proved to be an effective data augmentation technique, improving the generalization and robustness of deep learning models. TransU-Net, incorporating Transformer structures, demonstrated superior segmentation accuracy and IoU metrics, outperforming both U-Net and Attention-U-net. Our research advances rock image segmentation by combining the diffusion model with cutting-edge neural networks, reducing dependency on extensive expert data and boosting segmentation accuracy and robustness. TransU-Net sets a new standard in digital rock physics, paving the way for future geoscience and engineering breakthroughs.

# Introduction

Accurate analysis of rock pore structures plays a vital role in predicting their performance in various engineering applications, such as $CO_2$ storage, geothermal energy extraction, and radioactive waste disposal. While experimental techniques like nuclear magnetic resonance can provide such insights, they come with significant drawbacks, including high costs and lengthy processes. In contrast, digital rock physics (DRP) presents a more efficient alternative, especially when dealing with low-permeability rocks where conventional experiments are slow.

DRP involves the utilization of X-ray technology to scan rock samples and generate CT images. These images are then used to characterize pore structures and estimate petrophysical parameters. The significance of DRP lies in its ability to provide essential insights into the microscopic properties of porous rocks, including porosity and permeability, as well as their interactions with various fluids. Such insights are fundamental in critical rock engineering applications, such as carbon dioxide sequestration, enhanced oil and gas recovery, and the management of underground water resources [1]. To perform computations related to petrophysical properties, it becomes essential to divide digital rock images into distinct components (such as pores and rock matrix). This segmentation process can range from the basic separation into rock skeleton and pores, to more intricate divisions involving various phases. Image segmentation entails the categorization of pixels (2D) or voxels (3D) into specific classes, such as pores, minerals, fractures, and so forth. The accuracy of this segmentation procedure holds significant importance as it directly impacts the precise characterization of subsequent physical properties.

While DRP proves to be effective, it faces several challenges, including the need to find a delicate equilibrium between resolution and field-of-view, mitigating the influence of noise and artifacts, and maintaining a high segmentation accuracy. Present micro-CT technologies have resolution constraints, often resulting in digital images with a significant amount of sub-resolution porosity, where the pore sizes are smaller than the image's resolution. Consequently, there have been research efforts to improve the resolution of CT images using super-resolution techniques, such as diffusion model [2]. Smal et al. [3] proposed a novel algorithm that allows localization and quantification of sub-resolution porosity. Accurate segmentation of various digital rock images, like CT scans, SEM images, and thin sections, is vital for accurately analysing physical properties.

Classical segmentation methods encompass techniques like the threshold method [4] (including the maximum interclass variance method [5]), cluster analysis and boundary detection [6]. However, these traditional methods are susceptible to notable errors stemming from image noise, artifacts, and user bias. This susceptibility introduces uncertainties into the segmentation outcomes [7-9]. For instance, threshold methods often struggle to automatically distinguish phases with similar colours and intensities in rock CT images, such as the challenge posed by the similar attenuation characteristics between quartz and feldspar [10]. To address the effect of noise, researchers commonly employ noise reduction techniques like median filtering. Some have even adopted advanced methods, such as the dual filtering approach, involving unsharp masking and median filtering, prior to applying regular thresholding segmentation methods [11]. Phan et al. [1] have developed a tool designed to fully automate the segmentation process in a single step, eliminating the need for additional image processing steps like noise reduction or artifact removal. Given the typically low contrast and high noise in rock CT scans, initial image enhancement and denoising steps are crucial prerequisites for segmentation.

Figure 1 illustrates the impact of noise on the visibility of pore structures and segmentation accuracy. It compares the results of k-means clustering on raw, noisy data versus denoised carbonate CT images. To eliminate the randomly distributed noise, deep-learning-based denoising method like SCUNet is employed. Figure 2 provides a comparative assessment of k-means clustering and Gaussian mixture model (GMM) segmentation, highlighting how ring artifacts can distort segmentation results. This comparison underscores the superior capability of k-means clustering in handling ring artifacts, outperforming the Gaussian mixture model in such scenarios. Nevertheless, both methods often require substantial manual intervention and quality control.

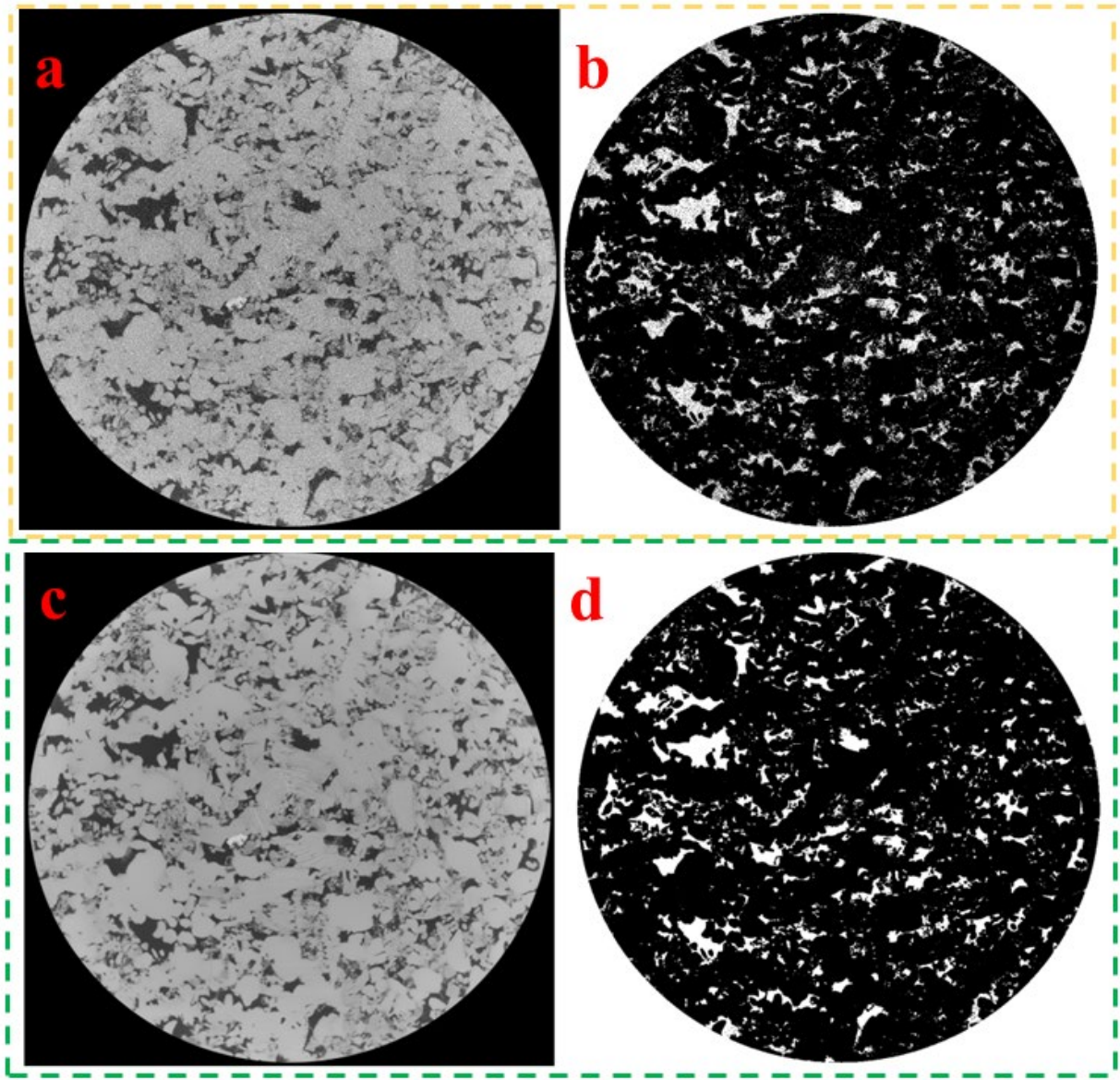

Figure 1: Comparative analysis of rock CT image segmentation using k-means clustering (a) raw noisy carbonate CT image (b) Binary segmentation result with k-means on raw noisy image (c) denoised carbonate CT image using the Swin-Conv-UNet (SCUNet) denoising network (d) Binary segmentation result with k-means on denoised image

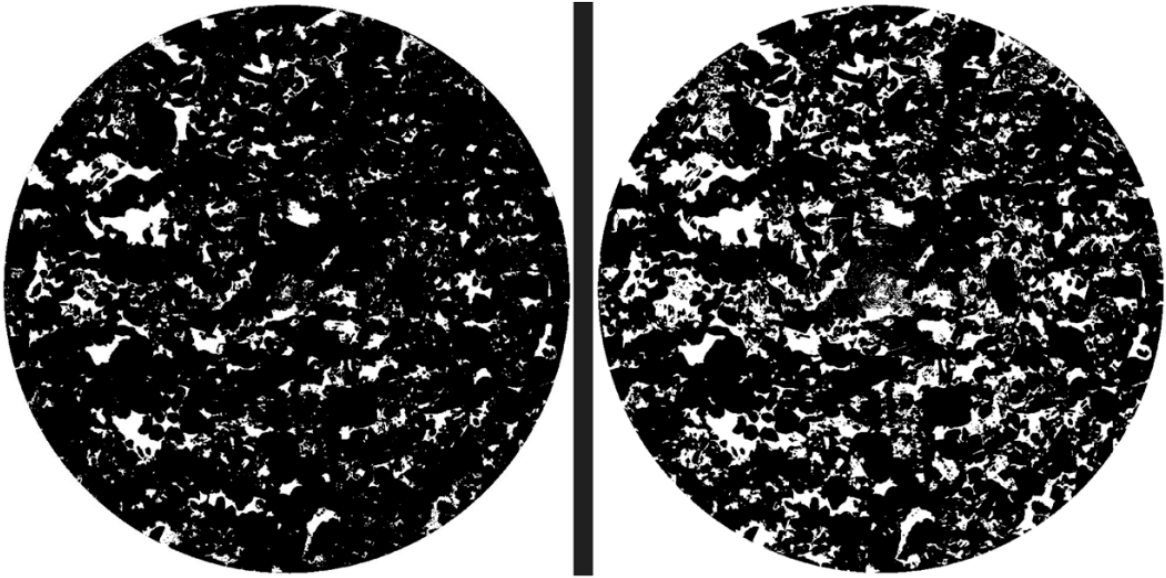

Fig. 2 K-means (left) vs Gaussian Mixture Model (GMM) (right) clustering in rock CT image segmentation.

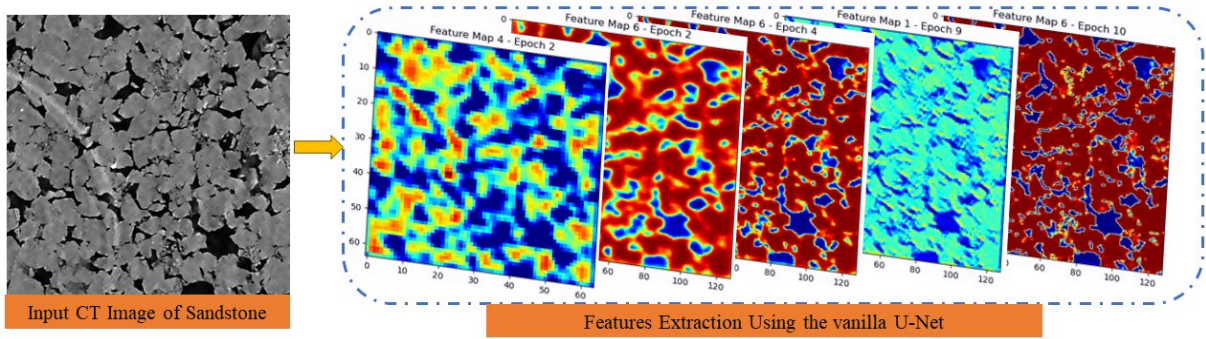

Fig. 3 Feature extraction using the vanilla U-Net with a CNN backbone.

Despite many years of intensive research in digital rock image segmentation, it continues to be a challenging issue largely because of uncertainties associated with segmentation, like the partial volume effect [8]. Thresholding, commonly used in image segmentation for its straightforward approach, often struggles with accurately extracting features, particularly when grain overlaps, or blurred object boundaries are present. Due to its subjectivity, the threshold segmentation hardly ensures the uniqueness of the segmentation results.

In contrast, the watershed algorithm excels in identifying overlapping nuclei structures, making it a popular choice for analysing pores, throats, and grain distributions in porous media. This algorithm operates by flooding local minima (representing dark spots like pores or organic matter) and local maxima (indicating bright spots or mineral matter) with distinct virtual markers. The process continues until only the peaks (pixels that act as barriers between the minima and maxima) remain visible, effectively submerging other features. This method creates more precise segmentation than intensity-based thresholding, as the barriers delineate the segmented areas. Recent advancements in the watershed algorithm, such as adaptive or marker-based variations, have further improved its ability to maintain edge definition and more accurately model contact points between different objects, enhancing its overall performance compared to traditional thresholding techniques [12]. However, digital rock image segmentation using the watershed method typically leads to over-segmentation problems, where pores are overestimated [13].

In comparison to machine-learning or deep-learning (ML/DL) techniques, traditional algorithms such as thresholding and watershed algorithm tend to be less effective, especially when dealing with images that have high levels of noise [9]. However, to train a segmentation model using the ML/DL, techniques, it is of vital importance to construct a substantial dataset, which is both time-intensive and laborious. Although there are lots of public datasets available for benchmarking various segmentation methods in the context of natural scenery images, the situation in the digital rock community differs significantly. Firstly, only a limited number of public datasets exist for this community (Table 1). Secondly, most of the available datasets often contain only raw images instead of raw image-ground truth segmentation pairs. This is because precise labelling demands the expertise of geoscientists who must engage in manual or semi-automatic segmentation.

Table 1. Public datasets in the digital rock community

| No. | Public digital rock image datasets name | Corresponding website |
|---|---|---|
| 1 | Digital Rocks Portal | https://www.digitalrocksportal.org/ |
| 2 | Micro-XRCT datasets for high-porosity volcanic rock | https://darus.uni-stuttgart.de/dataset.xhtml?persistentId=doi:10.18419/darus-680 |
| 3 | CT data for interparticle contact detection analysis in spheroidal granular packings | https://zenodo.org/records/5806270 |
| 4 | Edge segmentation of grains in thin-section petrographic images | https://github.com/ELOESZHANG/ECPN |
| 5 | High-resolution time-resolved synchrotron X-ray CT datasets of drainage and imbibition in carbonate rocks at reservoir pressure conditions | https://www2.bgs.ac.uk/ngdc/accessions/index.html#item112690 |
| 6 | Micro-XRCT data set of Carrara marble with artificially created crack network | https://darus.uni-stuttgart.de/dataset.xhtml?persistentId=doi:10.18419/darus-682 |

While Convolutional neural networks (CNNs) offer advantages such as translation invariance and localized receptive fields, they lack a holistic understanding of images, concentrating solely on the presence of features without comprehending the structural relationships between these features, as illustrated in Figure 3. Furthermore, CNNs are computationally demanding, especially in the context of 3D or 4D segmentation, rendering them unsuitable for usage on low-performance computing systems. Another notable limitation of CNN-based segmentation is its domain-specific design, which hampers scalability and application in domains beyond their original scope. In essence, CNNs operate based on pixel arrays, with the significance of each pixel contingent on its surrounding context. In simpler terms, the efficacy of CNNs in segmentation relies on the availability of training sets containing samples with similar visual and structural characteristics. For example, a CNN model proficient in segmenting sandstone images may not perform as effectively on carbonate images. Consequently, data augmentation techniques are often employed to improve the generalization capabilities of these models.

Recently, the classical U-Net, which adopts an encoder-decoder structure (Fig. 4) proposed by Ronneberger et al. [14] , has gained significant popularity, primarily due to its impressive segmentation accuracy. In the initial stages, a 2D CT image input undergoes a downsampling process within the encoder, which involves convolutional blocks, resulting in the creation of feature representations. These representations are subsequently upscaled to their original dimensions using deconvolutional blocks within the decoder, ultimately producing what is commonly referred to as a segmentation mask. The U-Net model incorporates skip-connections between corresponding blocks of the encoder and decoder, strategically designed to minimize information loss. This innovative approach significantly elevates the quality of segmentation outcomes. The U-Net architecture has found successful applications in the development of surrogate models for pore-scale image segmentation, demonstrating exceptional performance when evaluated on previously unseen images [15-18].

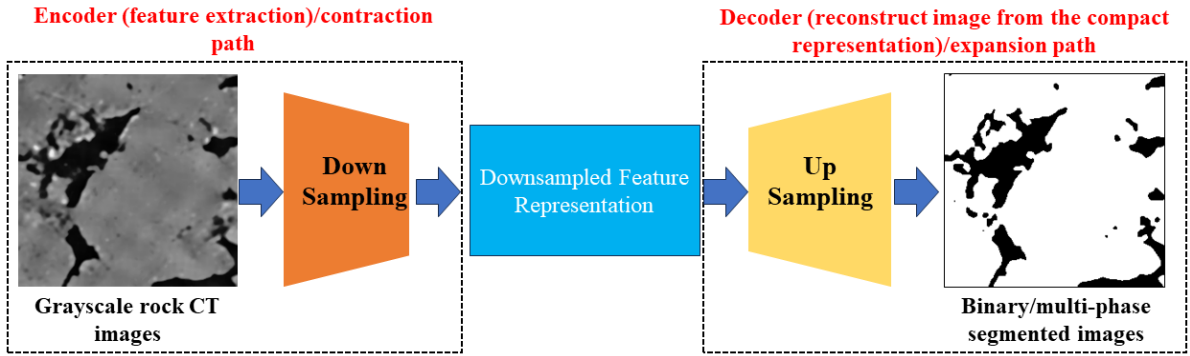

Fig. 4 General structure of the encoder-decoder architecture for digital rock image segmentation.

A series of variants of U-Net structures, including U-Net++ [19, 20], ResU-Net [21, 22], Attention U-Net [23], TransU-Net [24] has been developed to further improves the segmentation performance given the success of the vanilla U-Net model. The integration of attention mechanisms has garnered significant interest and demonstrated remarkable results, outperforming traditional CNNs in the field of image segmentation. Transformers, with their ability to grasp global context, have been adopted by researchers as encoders in innovative models like TransUNet. This approach integrates the concept of the Vision Transformer (ViT) into the encoder section of the U-Net, enhancing its capabilities. Transformers are believed to surpass traditional neural networks, such as CNNs, in both accuracy and computational efficiency. ViT [25] is emerging as a strong alternative to CNN for handling extensive datasets, owing to its greater model capacity and improved generalization, primarily stemming from its absence of inductive bias. However, employing ViT directly for digital rock image segmentation presents certain challenges due to this very absence of inductive bias. Effective training necessitates a substantial quantity of meticulously annotated digital rock images.

In this research, our primary goal is to tackle the challenge of precisely segmenting grayscale images, particularly in the context of supervised learning-based segmentation methods like U-Net, Attention-U-net, and TransU-Net. However, these methods often face limitations due to a shortage of labeled data. To overcome this hurdle, we utilized the diffusion model to generate an extensive dataset of CT/SEM images and binary segmentation pairs. In addition, this study marks another significant milestone in the realm of digital rock physics, as we harness the capabilities of TransU-Net, incorporating Transformer structures, to establish a substantial segmentation performance benchmark.

## Methods

The accurate segmentation of digital rocks into categories like pore and rock matrix in 2D or 3D images remains a formidable challenge. Traditional techniques such as global thresholding and watershed segmentation are prone to user bias and sensitivity to image noise and artifacts. An improved segmentation model should ideally require minimal to no user intervention, thereby reducing the influence of user subjectivity. With advancements in computational power and the growing size of digital rock datasets, machine learning has gained increasing prominence as a solution for these challenges. However, the success of deep learning in semantic segmentation depends on the large number of training data with accurate pixel-level human annotations, which are prohibitively expensive and time-consuming to collect, especially for digital rock communities.

Data augmentation and expanding the dataset can improve the segmentation performance. To achieve a robust model, it's essential to ensure the precision of ground truth segmentation masks and to include a diverse range of input images, varying in rock type, image quality, and noise levels [15]. CNN-based segmentation methods exhibit promise but exhibit a notable drop in performance when dealing with limited labeled data. Specifically, CNN models trained with cross-entropy loss are susceptible to overfitting when confronted with a limited quantity of labeled data [26].

Several factors contribute to the limited availability of labeled data in the field of digital rock physics. A significant challenge is the need for specialized knowledge to accurately interpret complex images like CT scans, SEM images, and thin sections. This expertise is crucial, especially for tasks such as multi-mineral segmentation where distinguishing between different minerals requires a high degree of skill and experience, often beyond the capabilities of many researchers. Furthermore, the diverse nature of rocks, encompassing various types like metamorphic, sedimentary, and igneous, each with their distinct heterogeneous properties, adds to the complexity and time required for data labelling. The field also suffers from relatively lower public interest and funding, especially when compared to areas like medical imaging or autonomous vehicles, leading to fewer resources for comprehensive data collection and labelling. Lastly, the specific goals of segmentation vary among researchers, with some focusing on pore structures, others on fractures, or grain boundaries, which further complicates the creation of a universally applicable labeled dataset.

To address the issue of limited labeled data in digital rock physics, several approaches have been employed. One option is to use semi-supervised or unsupervised learning methods. However, the diverse and unique nature of each rock sample presents a significant challenge for these methods to effectively generalize across various rock types and structures. Additionally, the complex structures and properties of rocks hinder the ability of unsupervised and semi-supervised methods to accurately identify and classify features without a considerable amount of labeled data. In contrast, data augmentation is commonly used as a strategy to enhance the training of robust models. This technique can expand the available dataset and improve model performance, even with limited labeled data. The traditional and widely favoured approach to data augmentation involves applying basic transformations such as rotations and flips to create new images from existing ones. Yet, this method often falls short in introducing significant diversity, particularly along crucial semantic dimensions inherent in the input data. To enhance the effectiveness of augmentation, more sophisticated techniques might be necessary to capture the full range of variability and intricacies present in the original dataset, ensuring a more comprehensive and diverse set of augmented images [27].

Table 2 presents a comparison of the benefits of using diffusion models versus conventional methods for data augmentation. To summarize, although traditional data augmentation techniques are straightforward and commonly applied, diffusion models provide a more advanced and potentially more efficient approach for data enhancement. This is particularly relevant in situations where maintaining realism, applying continuous transformations, and preserving semantic content are essential. Likewise, Figure 5 offers a comparative analysis of conventional data augmentation and generative model-based augmentation within digital rock imaging studies. Generative models are adept at creating a vast number of images while ensuring high resolution and quality. They also provide the ability to meticulously tailor the content of the generated images to meet specific research needs. This approach serves as an effective remedy for the prevalent issue of data limitation by furnishing a more expansive collection of data. Furthermore, it refines the augmentation methodology by producing a range of synthetic images that are both diverse and realistic, thereby closely emulating the actual diversity present in rock CT imagery.

Table 2. Comparison between the conventional data augmentation and diffusion model

| No. | Conventional data augmentation | Generative model (diffusion model as an example) |
|---|---|---|
| 1 | Often relies on basic image manipulations, which may not fully capture the range of variations found in | Diffusion models can generate more realistic variations of digital rock images by simulating a natural process of diffusion. This can be particularly beneficial in creating a |

| | actual digital rock images. | dataset that closely resembles real-world digital rock images, which is essential for training robust models. |
|---|---|---|
| 2 | Typically involve discrete, set transformations like 90-degree rotations or mirroring, which may not encompass the subtler changes that occur in natural environments. | Diffusion models can produce a continuum of transformations. This continuous nature can help in generating a more diverse dataset, which could potentially lead to better generalization in the trained models. |
| 3 | Flipping or rotating images excessively or improperly can lead to a loss of important semantic information. | Diffusion models can be designed to preserve the essential semantic characteristics of the data while still introducing variability. |
| 4 | Often lack a solid theoretical underpinning, making systematic improvements challenging. | Diffusion models are based on an underlying physical or mathematical model, which can provide a principled way of augmenting data. This model-based approach can also allow for better control and understanding of the augmentation process. |
| 5 | Fixed, pre-defined transformations. | Diffusion models can be learned from the data itself, adapting the augmentation process to the specific characteristics of the data. |

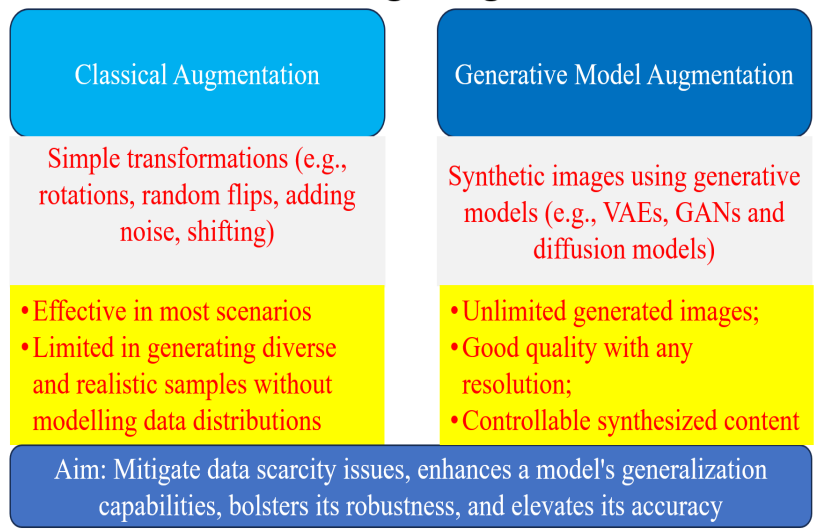

Fig. 5 Comparison of traditional and generative model augmentation methods.

Diffusion model [28, 29] can generate samples via a Markov chain with learned Gaussaian transitions starting from an initial noise distribution. Diffusion models are trained to grasp an unseen fundamental distribution via a dual-phase method: (i) initially, they progressively introduce random noise (forward process); (ii) subsequently, they reverse this process by eliminating noise for the purpose of sampling (reverse process) (Fig. 6). To accomplish this, diffusion models utilize a parallel approach known as score-based generative models (SGMs). SGMs execute the two-stage process described above through continuous dynamics, which are represented by a set of interconnected stochastic differential equations (SDEs) [30]. Apart from being a deep generative model, some researchers also adopt diffusion models for semantic segmentation [31] and super-resolution [2].

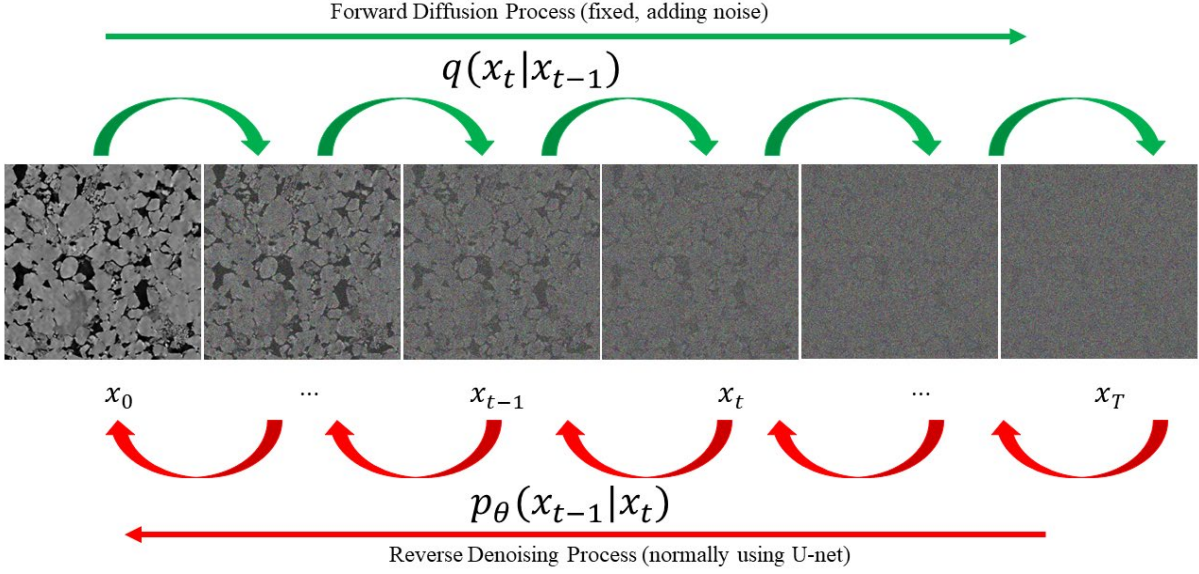

Fig. 6 Two stages of diffusion model

In the forward noising process denoted by $q$, we establish a Markov chain that iteratively introduces Gaussian noise into an image over a predetermined number of steps, $T$. This process adheres to a specified variance schedule characterized by an increasing sequence of variance parameters $\beta_0, \beta_1, \cdots \beta_t$, where $\beta_0 < \beta_1 < \cdots < \beta_t$. Consequently, we proceed to train a neural network model with the objective of sequentially reversing the noise introduced at each step. The model's training is contingent on the fact that the noise increments are sufficiently small, enabling the model to effectively learn the noise cancellation at every timestep.

Figure 6 illustrates a sequential relationship where $x_1$ is derived solely from $x_0$, $x_2$ from $x_1$, and this progression continues until we arrive at $x_T$. Here, $x_0$ represents the clean digital rock image. Progressing through the Markov chain, the image incrementally accrues noise, culminating in the state $x_T$, which is the noisiest version. Starting with a data point $x_0$ that follows the true data distribution $x_0 \sim q(x)$, the forward diffusion process introduces a small quantity of Gaussian noise at each step, resulting in a series of increasingly noisy images $x_1, x_2, \cdots, x_T$. As the step $t$ increases, the original image $x_0$ progressively loses its distinctive features. In the limit, as $T$ approaches infinity, $x_T$ becomes indistinguishable from an isotropic Gaussian distribution.

Because small amounts of random noise have been added progressively to the input digital rock image during the forward diffusion process and it follows the rule of Markov chain,

$$x_t = f(x_{t-1}) \tag{1}$$

To be more specific,

$$x_t = \sqrt{1-\beta_t}\, x_{t-1} + \sqrt{\beta_t}\, Z_t \tag{2}$$

$$Z_t \sim N(0, I) \tag{3}$$

The forward process variances $\beta_t$ can be learned by reparameterization, in order to obtain any $x_t$ from the $x_0$ state, we assign:

$$1 - \beta_t = \alpha_t \tag{4}$$

Therefore, Eq. (2) can be rewritten as:

$$x_t = \sqrt{\alpha_t}\, x_{t-1} + \sqrt{1-\alpha_t}\, Z_t \tag{5}$$

Similarly,

$$x_{t-1} = \sqrt{\alpha_{t-1}}x_{t-2} + \sqrt{1-\alpha_{t-1}}Z_{t-1} \tag{6}$$

We can combine Eq. (5) and (6):

$$x_t = \sqrt{\alpha_t}(\sqrt{\alpha_{t-1}}x_{t-2} + \sqrt{1-\alpha_{t-1}}Z_{t-1}) + \sqrt{1-\alpha_t}Z_t \tag{7}$$

Therefore, we can obtain the following equation:

$$x_t = \sqrt{\alpha_t\alpha_{t-1}}x_{t-2} + \sqrt{\alpha_t-\alpha_t\alpha_{t-1}}Z_{t-1}) + \sqrt{1-\alpha_t}Z_t \tag{8}$$

Because both $Z_t$ and $Z_{t-1}$ obey the Gaussian distribution, namely, $Z_t \sim N(0,I)$ and $Z_{t-1} \sim N(0,I)$. When we merge two Gaussian's with different variance, $N(0, \sigma_1^2 I)$ and $N(0, \sigma_2^2 I)$, the new distribution is:

$$N(0, (\sigma_1^2 + \sigma_2^2)I)$$

Because in Eq. (8), both $\sqrt{\alpha_t - \alpha_t\alpha_{t-1}}Z_{t-1}$ and $\sqrt{1-\alpha_t}Z_t$ follows the Gaussian distribution, namely,

$$\sqrt{\alpha_t - \alpha_t\alpha_{t-1}}Z_{t-1} \sim N(0, \alpha_t - \alpha_t\alpha_{t-1}) \tag{9}$$

$$\sqrt{1-\alpha_t}Z_t \sim N(0, 1-\alpha_t) \tag{10}$$

Since the mean value is 0 and the sum of $\sqrt{\alpha_t - \alpha_t\alpha_{t-1}}Z_{t-1}$ and $\sqrt{1-\alpha_t}Z_t$ also follows the Gaussian distribution, namely,

$$\sqrt{\alpha_t - \alpha_t\alpha_{t-1}}Z_{t-1}) + \sqrt{1-\alpha_t}Z_t \sim N(0, 1-\alpha_t\alpha_{t-1}) \tag{11}$$

Therefore, equation (8) can be rewritten as follows:

$$x_t = \sqrt{\alpha_t\alpha_{t-1}}x_{t-2} + \sqrt{1-\alpha_t\alpha_{t-1}}Z \tag{12}$$

where

$$Z \sim N(0,I) \tag{13}$$

Similarly,

$$x_t = \sqrt{\alpha_t\alpha_{t-1}\cdots\alpha_1}x_0 + \sqrt{1-\alpha_t\alpha_{t-1}\cdots\alpha_1}Z \tag{14}$$

From Eq. (14), we can obtain $x_0$

$$x_0 = \frac{x_t - \sqrt{1-\alpha_t\alpha_{t-1}\cdots\alpha_1}Z}{\sqrt{\alpha_t\alpha_{t-1}\cdots\alpha_1}} \tag{15}$$

For the reverse diffusion process, $\tilde{Z}$ can be obtained from the U-Net instead of $Z$, namely,

$$x_0 = \frac{x_t - \sqrt{1-\alpha_t\alpha_{t-1}\cdots\alpha_1}\tilde{Z}}{\sqrt{\alpha_t\alpha_{t-1}\cdots\alpha_1}} \tag{16}$$

When using U-Net, for a particular $x_t$, we can obtain the $\tilde{Z}$, which can be described as $\tilde{Z} = UNet(x_t, t)$. To be more specific, how to gain a function as follows is the most significant issue for the reverse diffusion process.

$$x_{t-1} = f(x_t, \tilde{Z}) \tag{17}$$

Fig. 7 illustrates the workflow adopted in this study for training U-net based segmentation models, enhanced by generative model augmentation techniques. Initially, a limited set of real digital rock images, along with their corresponding binary segmentation pairs, is provided. These are used to leverage a diffusion model, which is then employed to generate an extensive collection of synthetic digital rock images and their binary segmentation pairs. Subsequently, this synthesized dataset, in conjunction with the original smaller dataset of real image pairs, is utilized in the training of U-Net and its variants, including the attention U-Net and TransU-Net. This approach aims to enhance the models' generalization capabilities, robustness, and prediction accuracy.

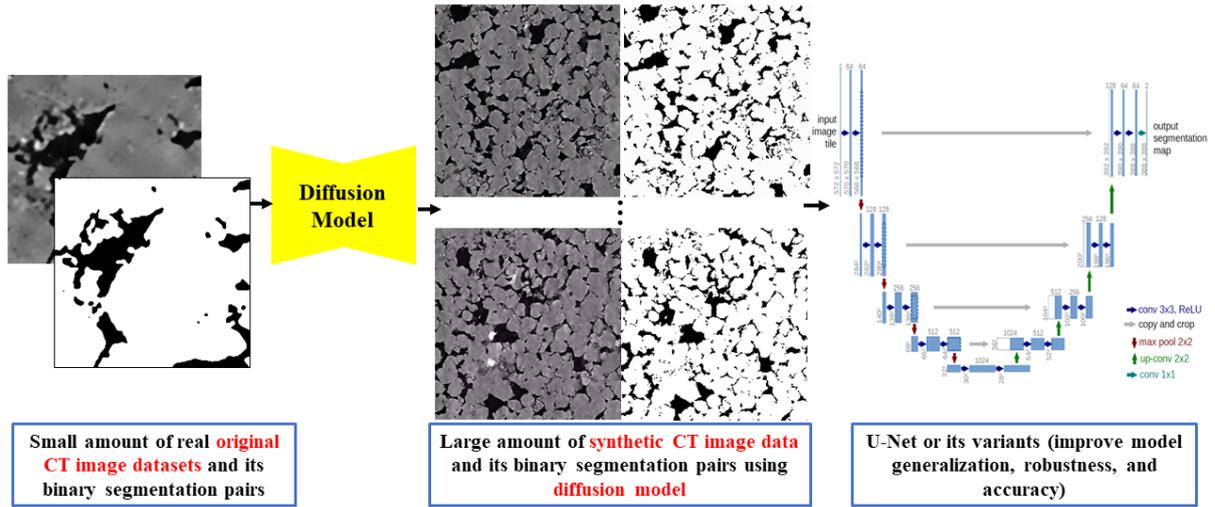

Fig. 7. Workflow to train U-net based segmentation models using generative model augmentation techniques.

In this paper, we harnessed the capabilities of the U-Net architecture within the PyTorch framework to address the challenges of image segmentation. The U-Net is notably proficient for this purpose, given its symmetrical encoder-decoder configuration. The encoder progressively compresses the input image into a dense feature representation, which the decoder subsequently expands to generate a detailed segmentation mask. This process is enhanced by skip-connections that facilitate the flow of information between corresponding layers in the encoder and decoder, thereby preserving crucial details for precise segmentation.

The most prominent networks in image segmentation CNNs are U-Net [14] and SegNet [32]. Sultana et al. [33] summarized the evolution of CNN based semantic segmentation models and they reviewed some popular state-of-the-art semantic segmentation models. U-Net effectively merges information from both low-level and high-level layers to produce dense predictions (pixel-level identification of the pores). Its proficiency in representation combined with efficient GPU memory utilization has established U-Net as the default, or de facto, method for image segmentation [34].

To address the limitations of U-Net in capturing long-range dependencies and learning global context—due to its inherently local convolution operations—TransUNet has been proposed as an innovative solution that merges the advantages of both Transformers and U-Net. Although CNNs have made notable strides in digital rock imaging over the last decade, their design is inherently limited, restricting their potential for enhanced performance. The small convolutional kernels (typically 3x3 or 5x5) used in most CNNs focus primarily on neighbouring areas, making them adept at detecting local patterns but less effective at recognizing and integrating broader spatial relationships that are essential for a thorough understanding of the contextual information in images. In contrast, transformers have shown superior results in image segmentation, offering scalability and greater resistance to corruption. Their ability to capture distant features, owing to the self-attention mechanism, gives transformers a distinct edge over CNNs. Furthermore, there is an expectation that transformers will continue to exhibit improved segmentation accuracy with increasing input sizes, as there is no evidence yet of their performance reaching a saturation point.

Unlike the traditional U-Net, TransUNet incorporates Transformer architectures which inherently possess global self-attention mechanisms. This offers a substantial improvement for handling long-range contextual information. In the TransUNet model, the Transformer component serves to encode tokenized image patches from a CNN feature map, thereby capturing global contexts effectively. Subsequently, the decoder up-samples these encoded features and integrates them with high-resolution CNN feature maps, ensuring precise localization. This synergy provides a more robust and versatile approach for image segmentation, especially when compared to vanilla U-Net [35].

## Data Augmentation training and segmentation results

Fig. 8 shows the reprehensive training dataset of sandstone, which is obtained from the digital rock portal (https://www.digitalrocksportal.org/). The training process of the diffusion models as well as using the U-net and its variants were run on the NVIDIA A100, the GPU memory per card is 80 GB.

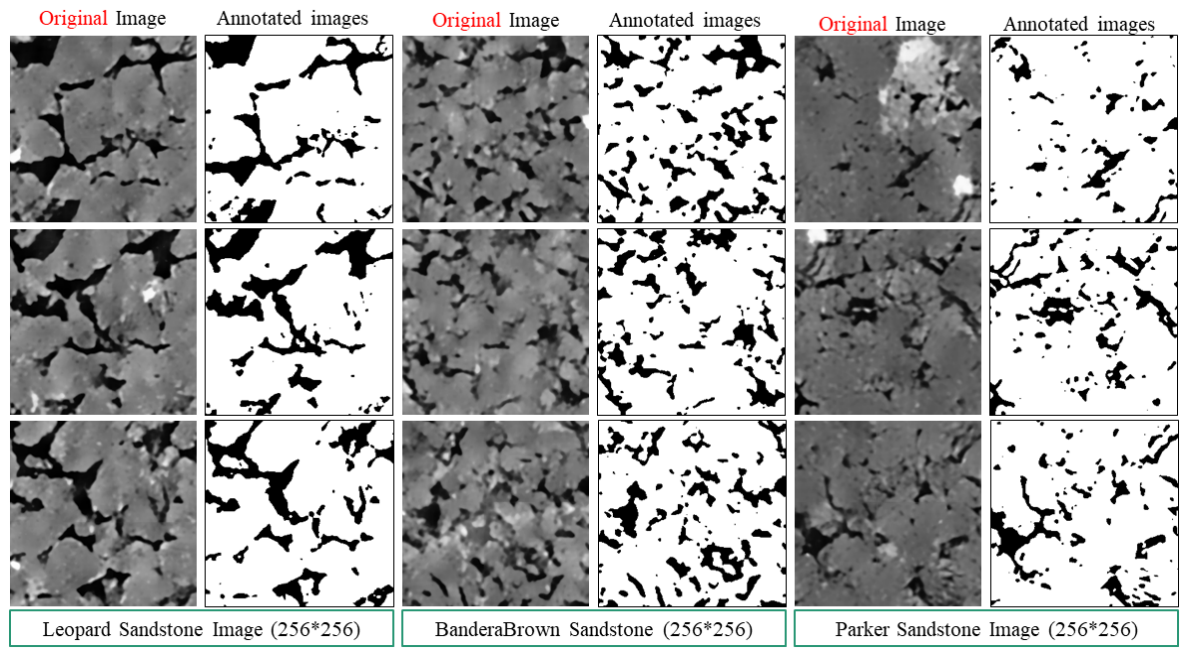

Fig. 8. Reprehensive training dataset of sandstone.

For digital rock image augmentation, we present a Python implementation of a guided diffusion model specifically designed for data augmentation in digital rock image processing. The framework, built on the PyTorch library, is configured to train on a diverse set of images. The key innovation lies in the model's ability to perform data augmentation through a controlled diffusion process, which is pivotal in generating new, varied data samples from existing datasets. The model's training loop involves iterative refinement of the model parameters to effectively learn the distribution of the input data. This is achieved through a stochastic process where the model gradually learns to denoise images, starting from random noise. This process inherently produces a range of intermediate images, each slightly different from the original, thus creating augmented data. The inclusion of a schedule sampler further enhances the model's ability to vary the diffusion process, contributing to the generation of diverse augmented images.

Furthermore, the flexibility of the model's architecture allows for adjustments in batch size, learning rate, and weight decay, among other parameters. This adaptability makes it suitable for various data augmentation needs in image processing tasks. Coupled with its ability to operate on multiple GPU configurations, the model presents a robust solution for enhancing the quality and variety of image datasets, a crucial aspect in training high-performance deep learning models in domains such as medical imaging, autonomous driving, or digital rock physics.

In short, the framework utilizes a custom dataset loader to handle various image datasets, equipped to process images with additional channels, such as segmentation masks. It initializes the diffusion model with user-defined parameters, offering flexibility in adapting to different data characteristics. The training loop is orchestrated using PyTorch, with provisions for distributed computing across multiple GPUs. A significant aspect of the implementation is the inclusion of a schedule sampler, which controls the diffusion process's timestep, a critical factor in the quality of generated images. This control mechanism is pivotal in tailoring the augmentation process to specific dataset requirements.

## Training

To compare the impact of training dataset size on the segmentation accuracy, we employed the binary cross-entropy with logits loss (BCEWithLogitsLoss), an apt choice for binary pixel-wise classification inherent to segmentation tasks. This function inherently integrates a sigmoid activation with a binary cross-entropy loss, providing a stable training phase particularly when the network's logits are highly positive or negative.

A suitable learning rate can improve the efficiency of the model training and get more accurate training results [36]. We utilized the Adam optimizer with a learning rate set at 0.001. This rate is widely accepted as an effective default that aids in achieving a balance between quick convergence and stability across various segmentation models and tasks. The U-Net model was constructed with an encoder for downsampling and a decoder for upsampling, each comprising convolutional blocks, connected by skip-connections. The model underwent training for a total of 350 epochs, a duration deemed sufficient for the network to adequately converge. To bolster the model's generalization, data augmentation techniques such as random rotations and horizontal flips were introduced during training. A custom dataset class managed the loading and preprocessing of images and masks, and DataLoaders were optimized for parallel processing using multiple workers.

Our methodology was applied to datasets of two distinct sizes to evaluate the impact of data volume on model performance. Performance metrics, including the Jaccard score, F1-score, and precision, were computed post-training to assess the model's efficacy in discriminating between segmented regions and the background. These metrics were chosen for their ability to offer a nuanced view of the model's performance across various segmentation thresholds. Our approach was meticulously crafted to fine-tune the U-Net model for segmentation tasks, with careful selection of loss functions and learning rates, complemented by a structured training and validation regimen. The model's performance was periodically evaluated during training, with checkpoints saved at points of minimum validation loss. This ensured that the most effective version of the model was retained for subsequent use or further refinement.

The comprehensive training and evaluation pipeline, executed on datasets of varying sizes, led to an exhaustive understanding of the model's performance, providing insights into the scalability and robustness of the U-Net architecture for image segmentation applications.

Figure 9 elucidates the efficacy of model complexity on the segmentation accuracy of a U-Net model trained on CT images. The segmentation results are strikingly similar to the ground truth, showcasing the model's proficiency. We examine three randomly selected CT images, their corresponding ground truths, and their segmented counterparts. The Intersection over Union (IoU) and accuracy metrics are calculated, revealing that the larger model, which employs synthetic images produced by the diffusion model, consistently outperforms the smaller model. This underscores the enhancement in segmentation performance attributable to data augmentation via synthetic image generation using the guided diffusion model.

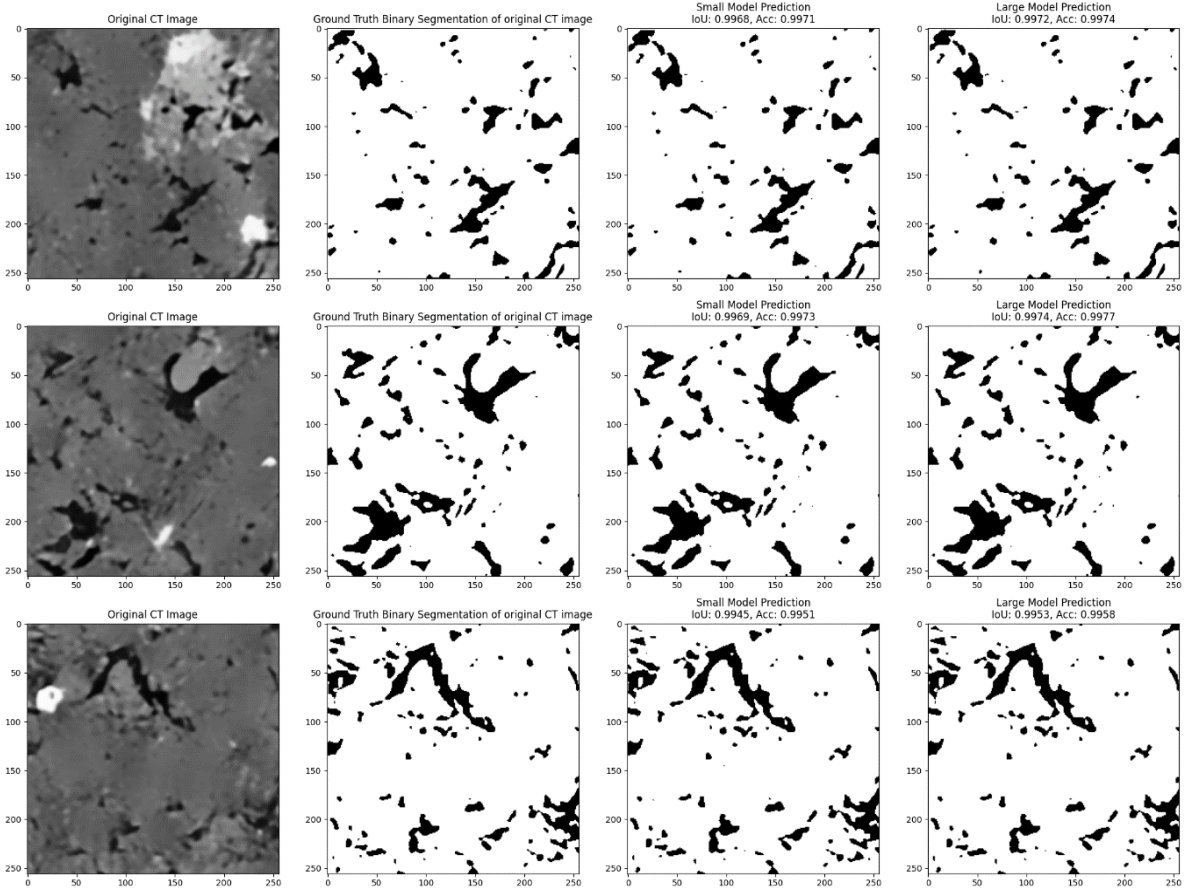

Fig. 9. Comparison of small and large model on three random CT images

In Figure 10, the generalization capability of the pre-trained model is scrutinized by applying it to novel SEM images not encountered during training. These images, along with their ground truth data, are extracted from the referenced study [37]. The model's predictions maintain a high IoU and accuracy, suggesting robust generalization. Notably, the predicted segmentation images present an improvement over the original ground truths in visual comparison, highlighting the model's potential in generalizing to diverse data types and suggesting that training with augmented datasets can lead to superior segmentation, even on previously unseen data.

This observation indicates that the inclusion of synthetic data in model training can significantly bolster the model's ability to accurately segment both familiar and novel images. This improvement is critical in applications where precise segmentation is paramount, such as medical diagnosis and digital rock physics, where the detailed differentiation of structures within images can be central to the analysis. The results from Figures 9 and 10 collectively suggest that data augmentation through diffusion models not only enhances the segmentation performance but also imparts the model with a strong ability to generalize across different image domains.

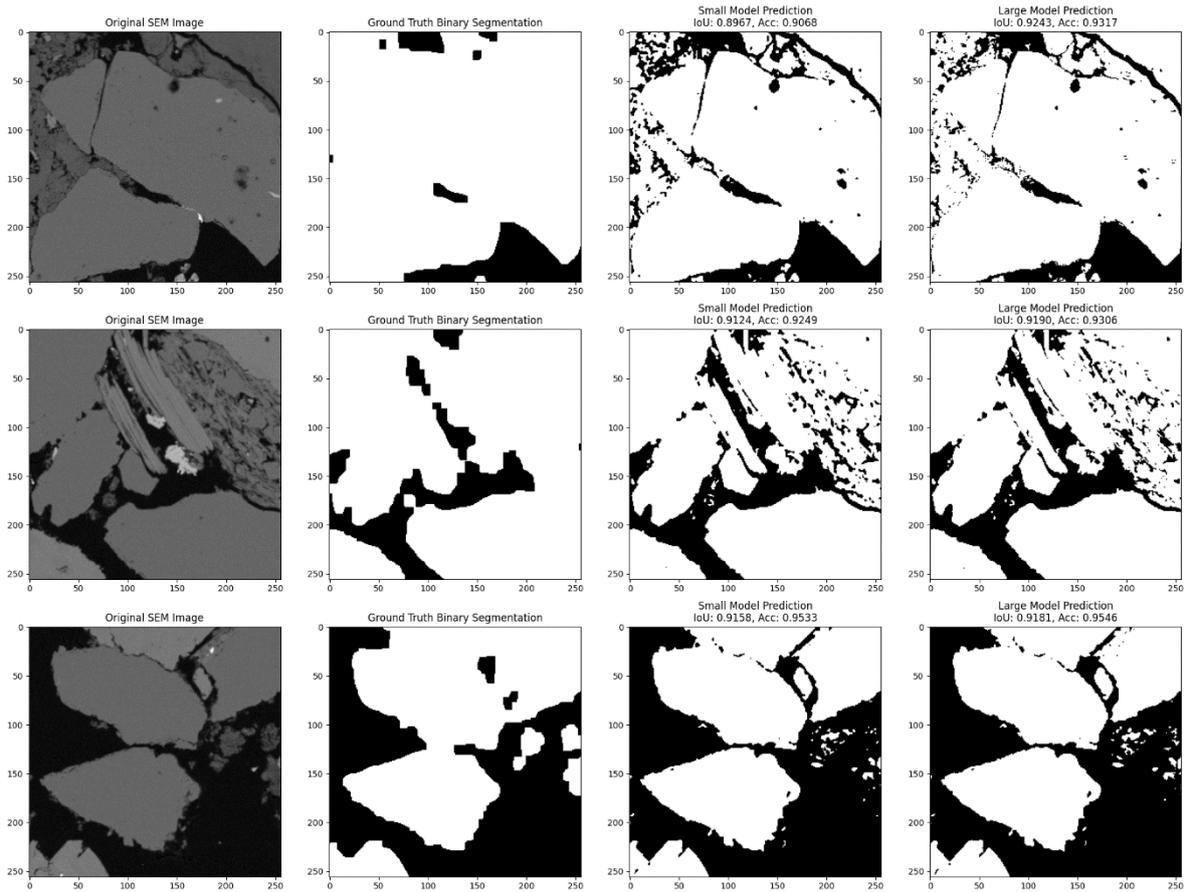

Fig. 10. Comparison of small and large model on three random SEM images

Accurate segmentation is of vital importance for the precise estimation of porosity, pore connectivity, fluid saturation, among other pore-scale properties [38]. Figure 11 presents a comparative analysis of morphological properties derived from segmented images, underscoring the significance of accurate segmentation for the precise quantification of pore-scale properties such as porosity, connectivity, and structural complexity. The large model's predictions more closely mirror the ground truth's porosity and connectivity, suggesting a more accurate representation of the physical space within the porous medium. This fidelity in representation is crucial for applications like fluid dynamic simulations in porous media, where such properties directly influence the simulation's accuracy.

The aspect ratio measurements from the large model also closely align with the ground truth, indicating a more accurate capture of the elongation of pores, which has implications for understanding flow pathways and material strength. Intriguingly, the fractal dimension—a measure of complexity over multiple scales—shows negligible differences across the ground truth, large model, and small model. This might suggest that both models are capturing the overarching complexity of the pore structure to a similar degree, despite the differences in other measured properties.

The data indicates that while the large model's augmented training dataset contributes to its superior performance in replicating porosity and connectivity, the inherent complexity of pore structures, as captured by the fractal dimension, seems to be less sensitive to the model size or the augmentation process. This nuance could be of particular interest in fields where the multiscale complexity of structures is of interest, such as in material science and geosciences. The slight discrepancy in fractal dimension values between the predictions and the ground truth also invites further investigation into the resolution and quality of the segmented images, as these factors can significantly influence fractal analysis.

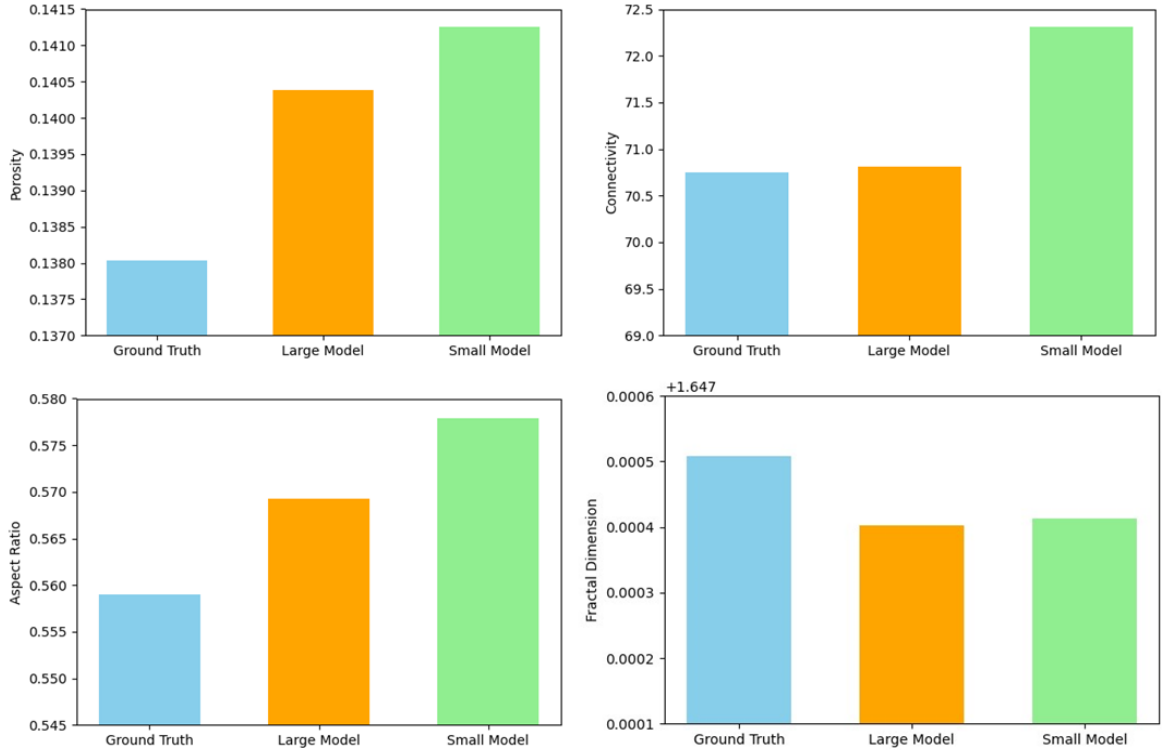

Fig. 11. Comparative Morphological Analysis of Segmented Images from U-Net Models of Varying Sizes

Figures 12 to 14 detail the performance metrics of semantic segmentation utilizing three variations of the U-Net architecture—standard U-Net, Attention U-Net, and TransU-Net. These graphs illustrate the training and validation loss, along with the Intersection over Union (IoU) across epochs, demonstrating the learning efficacy and validation accuracy of each model. The loss curves exhibit a typical convergence pattern, with the validation loss closely following the training loss, suggesting a good generalization capability without significant overfitting. The IoU curves, maintaining a high plateau, indicate a strong correspondence between the predicted segmentation and ground truth.

Figure 15 compares the segmentation results of the three models on unseen data, providing a visual assessment alongside quantitative IoU scores. The models exhibit remarkable segmentation capabilities, closely aligning with the ground truth binary images. Notably, the Attention U-Net and TransU-Net show a notable improvement in IoU scores compared to the standard U-Net, which can be attributed to their advanced mechanisms for capturing contextual relationships within the images. The Attention U-Net incorporates attention mechanisms to focus on relevant features for segmentation, while the TransU-Net utilizes transformer-based architectures known for capturing long-range dependencies, which is essential for understanding complex spatial structures in segmentation tasks.

The comparative analysis across Figures 12 to 15 demonstrates that incorporating attention mechanisms and transformer models into the U-Net architecture significantly enhances the model's performance, particularly in discerning intricate patterns necessary for accurate segmentation. This underscores the potential of advanced neural network architectures in improving the precision of automated image segmentation, which is critical in fields requiring detailed image analysis, such as medical imaging and digital rock physics.

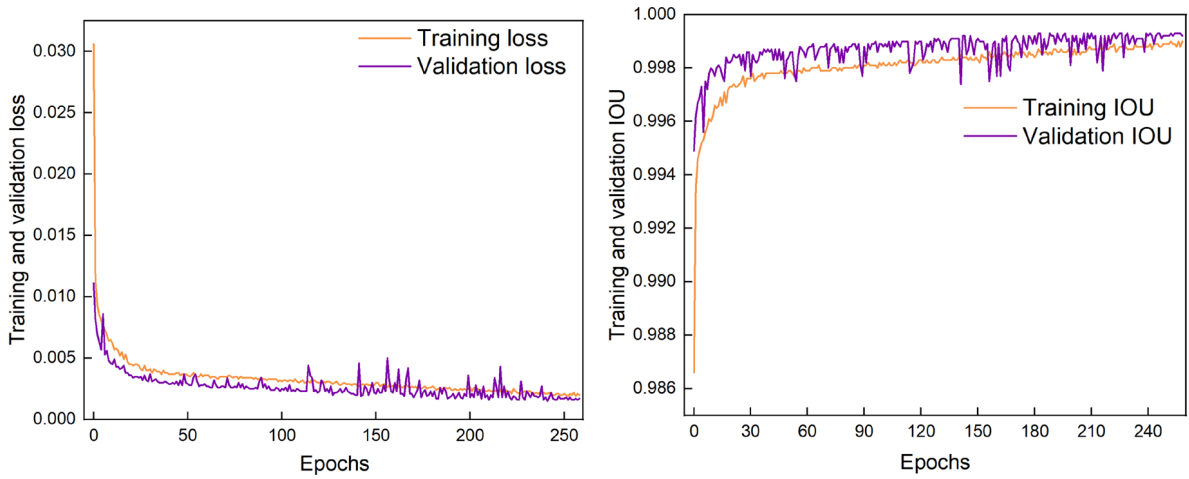

Fig. 12. Loss and IoU curves of the semantic segmentation using the U-net model.

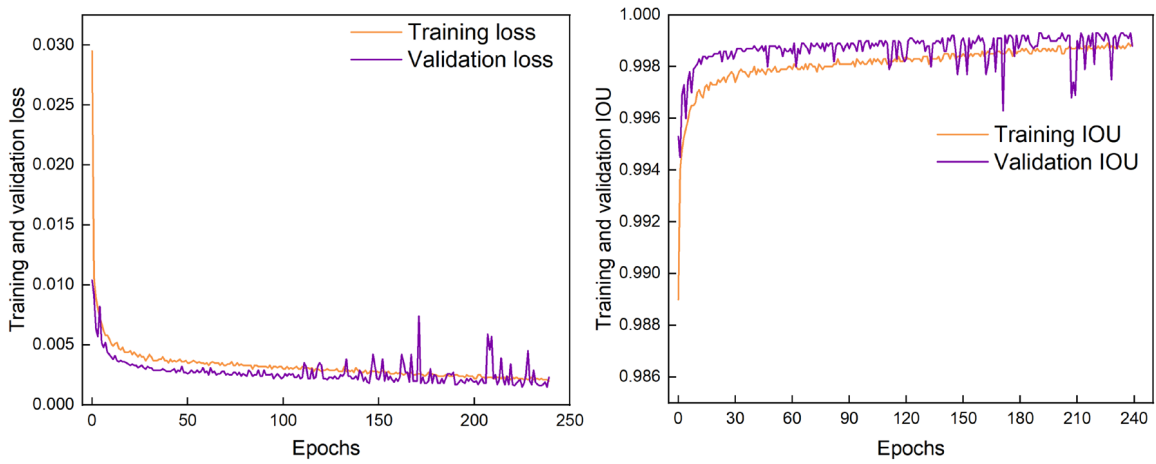

Fig. 13. Loss and IoU curves of the semantic segmentation using the Attention U-net model.

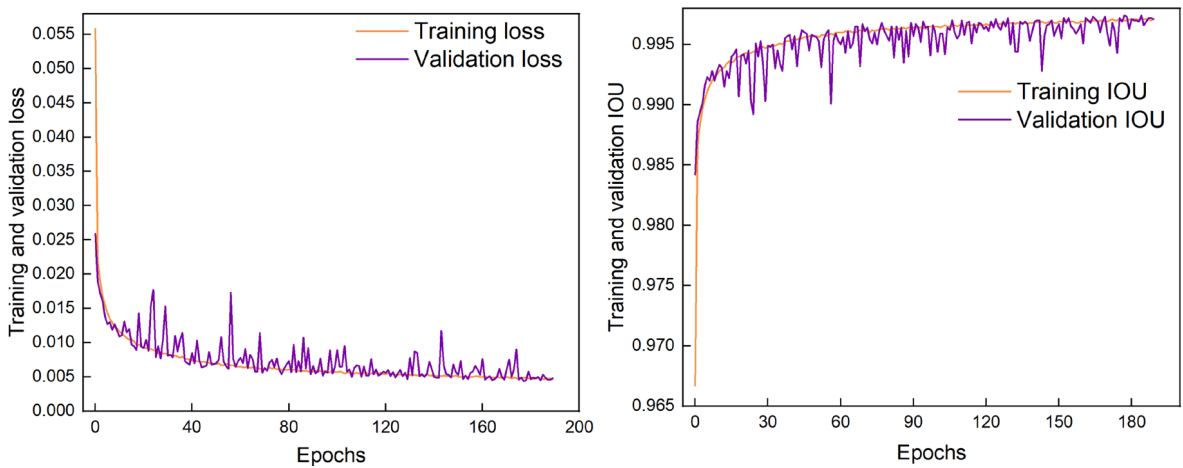

Fig. 14. Loss and IoU curves of the semantic segmentation using the TransU-net model.

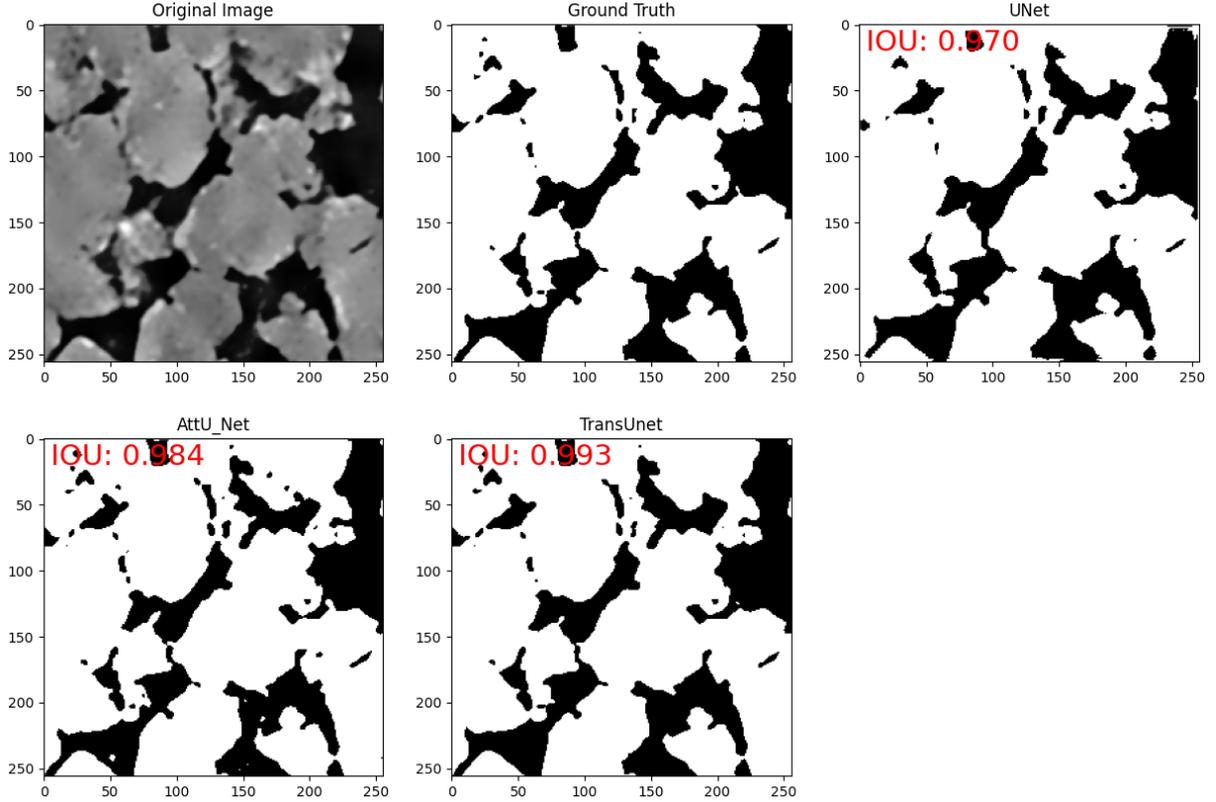

Fig. 15. Comparison between U-net, attention U-Net and TransUnet to compare with the Ground truth segmentation binary image.

Performance Evaluation:

As stated by [39], two main criteria are normally adopted to evaluate the performance of semantic segmentation: accuracy and computation complexity in terms of speed and memory requirements. The loss function is utilized to guide the network to learn meaningful predictions that are closed to the ground truth masks in terms of segmentation metrics during the training phase and a systematic study of the utility of loss functions in segmentation has been studied by Ma et al. [40]. The optimizer is utilized to minimize the loss function and generally, stochastic gradient descent method as well as its variant (e.g., Adam) are the most popular choices.

Intersection over Union (IoU) loss [41], similar to Dice loss, is also used to directly optimize the object category segmentation metric. It is defined by:

$$L_{IoU} = 1 - \frac{\sum_{c=1}^{C}\sum_{i=1}^{N} g_i^c s_i^c}{\sum_{c=1}^{C}\sum_{i=1}^{N}(g_i^c + s_i^c - g_i^c s_i^c)} \qquad (18)$$

$L_{IoU}$ represents the IoU loss, the number 1 signifies the maximum possible value of the IoU, indicating perfect overlap between the predicted and ground truth segmentation. $C$ denotes the number of classes in the segmentation task. In multi-class segmentation, each class will be evaluated separately. $N$ is the total number of pixels in the image or the batch of images being evaluated. $g_i^c$ indicates whether pixel $i$ actually belongs to class $c$ in the ground truth segmentation. If the pixel is part of the class, $g_i^c$ is 1; otherwise, it is 0. $s_i^c$ is the predicted probability that pixel $i$ belongs to class $c$. This value ranges from 0 to 1, with higher values indicating higher confidence in the prediction.

The IoU loss is calculated by taking one minus the sum over all classes of the intersection of the ground truth and the prediction (the sum of the element-wise product of $g_i^c$ and $s_i^c$) divided by the sum over all classes of the union of the ground truth and the prediction (the sum of $g_i^c$ plus $s_i^c$) minus their element-wise product).

In this study, the numerator quantifies the intersection between the predicted segmentation and the ground truth, whereas the denominator encompasses the union of both the predicted and actual segmentations. Subsequently, the Intersection over Union (IoU) loss is computed by deducting the ratio of the intersection to the union from one. This results in a metric that inversely correlates with the congruence between the predicted and ground truth segmentations; a diminution in the IoU loss signifies an enhancement in model performance.

Additionally, other performance metrics, such as accuracy, precision ($P$), recall ($R$), and the $F_1$ score, are also employed (refer to Figure 16). Within the context of Figure 16, True Positives (TP) represent the quantity of pixels accurately predicted as pore in alignment with the ground truth image. False Negatives (FN) denote the pixels erroneously predicted as rock matrix, whereas they are pore. False Positives (FP) are identified as the pixels incorrectly predicted as pore yet are in fact rock matrix. Lastly, True Negatives (TN) are those pixels correctly predicted as rock matrix, corroborating their actual state.

**Predicted Value**

|  | Pore | Rock Matrix |
|---|---|---|
| **Pore** | True Positive (TP) | False Negative (FN) |
| **Rock Matrix** | False Positive (FP) | True Negative (TN) |

(Actual Value on vertical axis)

Fig. 16 Definition of evaluation metric for segmentation

$$P = \frac{TP}{TP + FP} \quad (19)$$

$$R = \frac{TP}{TP + FN} \quad (20)$$

$$Accuracy = \frac{TP + TN}{FP + TP + TN + FN} \quad (21)$$

$$F_1 = 2 \times \frac{P * R}{P + R} \quad (22)$$

# Discussion

The research presented in the paper signifies a pivotal advancement in the field of digital rock physics (DRP), specifically in enhancing rock image segmentation through the integration of generative AI (diffusion model) and state-of-the-art neural networks. The study distinctly illustrates the limitations of traditional threshold segmentation and conventional CNNs in capturing the intricate details of rock microstructures. The introduction of advanced techniques, including the diffusion model for data augmentation and the use of sophisticated neural network architectures like U-Net, Attention-U-net, and TransUNet, marks a significant step forward.

The diffusion model's ability to generate an extensive dataset of CT/SEM and binary segmentation pairs from a limited dataset is particularly noteworthy. This approach addresses the challenge of the supervised nature of U-Net, which typically requires a substantial collection of expert-annotated samples, thus reducing labour intensity and susceptibility to human error.

TransU-Net's superior performance in terms of segmentation accuracy and Intersection over Union (IoU) metrics is a critical finding. This underscores the benefits of incorporating Transformer structures in neural networks for tasks requiring detailed image analysis, such as in DRP. The research also highlights the importance of data augmentation in enhancing the generalization and robustness of deep learning models, a factor crucial in the precise estimation of petrophysical parameters.

# Conclusion

This study heralds a pivotal advancement in the field of digital rock physics, tackling the longstanding issue of precisely segmenting digital rock images, such as CT and SEM scans. These images are indispensable for the construction of precise digital rock models, a cornerstone in understanding and analysing rock microstructures and their petrophysical properties. Historically, the domain has heavily relied on supervised learning methods, especially the U-Net architecture, but has been consistently hampered by the limited availability of labeled data pairs. Our research circumvents this challenge through the strategic fusion of generative AI with cutting-edge neural network architectures. A cornerstone of this innovative approach is the deployment of the diffusion model for data augmentation, markedly enhancing the generalization capabilities and robustness of the segmentation models. This not only diminishes the reliance on extensive, expert-curated datasets but also paves the way for more streamlined and efficient processes in digital rock analysis.

The introduction of TransU-Net in digital rock physics, incorporating advanced Transformer structures, is a standout achievement in this new landscape. It establishes new benchmarks in segmentation accuracy and Intersection over Union (IoU) metrics, propelling the capabilities of digital rock physics to unprecedented levels. The remarkable precision in rendering rock microstructures achieved by TransU-Net is crucial for the accurate assessment of petrophysical properties. This breakthrough sets a new standard for future research and applications in digital rock physics, promising significant contributions to the fields of geoscience, petroleum engineering, and beyond. Our findings not only contribute to the technological advancement of image segmentation but also open up new avenues for exploration and analysis in the digital representation of complex geological structures.

# Future Work

The upcoming phase of this research will be dedicated to fully exploiting the combined strengths of the CNN-Transformer architecture embodied in TransUNet. This approach synergizes the CNN's aptitude for capturing intricate, high-resolution spatial details with the Transformer's prowess in encoding comprehensive global context. Such integration positions TransUNet to achieve unprecedented performance in image segmentation.

Our future initiatives include: (1) broadening data spectrum: we aim to train our models on a more diverse array of digital rock images. This will include a wider range of rock types and imaging techniques, thereby enhancing the models' robustness and their adaptability to various geological contexts. (2) generalizing across varied rock types: recognizing the current models' limited efficacy mainly with homogeneous sandstone, we are committed to extending their proficiency to more heterogeneous rock types, such as carbonate and shale. This will involve thorough experimentation and refinement of data augmentation strategies and the U-Net segmentation framework, utilizing both CNN and transformer structures to assure comprehensive reliability and applicability across a spectrum of rock textures. (3) extensive testing on diverse samples: a key focus will be on rigorously evaluating these models against a broad range of previously unencountered objects and imaging methods. This is to further validate their generalization capabilities and to ascertain their precision in real-world scenarios. (4) incorporating semi or unsupervised learning approaches: we plan to explore and adopt semi or unsupervised deep learning methods for digital rock image segmentation, aiming to enhance model efficiency and reduce dependency on labeled data.

By pursuing these objectives, our mission is to drive forward the field of digital rock physics, pushing the limits of image segmentation accuracy and expanding the practical reach of these sophisticated techniques.

## Declaration of competing interest

The authors declare that they have no known competing financial interests or personal relationships that could have appeared to influence the work reported in this paper.

## Data availability

Data will be made available on request. The code will upload to the GitHub when the manuscript is accepted.

## Acknowledgement:

This work signifies a collaborative endeavour led by Prof. Shuyu Sun and Prof. Bicheng Yan, generously supported by Saudi Aramco. For computer time, this research used the resources of the Supercomputing Laboratory at King Abdullah University of Science & Technology (KAUST) in Thuwal, Saudi Arabia. We thank anonymous reviewers for their specific comments and instructive suggestions. Thank the Digital Rocks Portal for providing the open source data.

## Supplementary materials:

The similar attenuation characteristics between quartz and feldspar (readers may refer to https://physics.nist.gov/PhysRefData/FFast/html/form.html to looking for the X-ray attenuation factors for different minerals that share almost the same density).

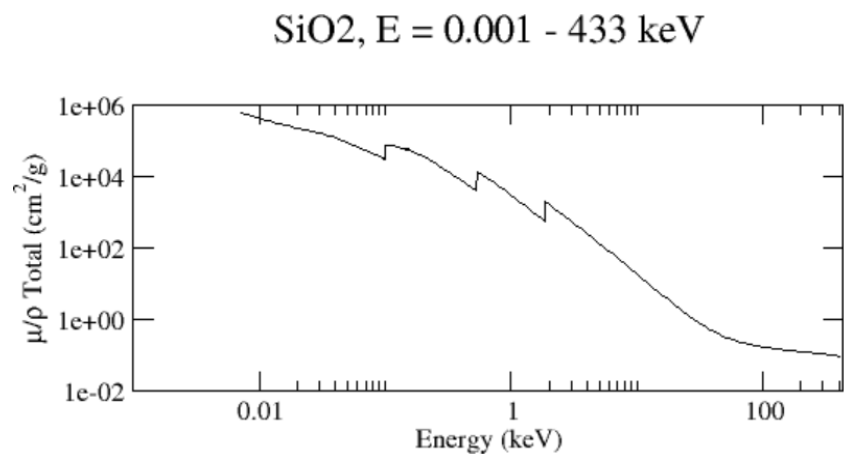

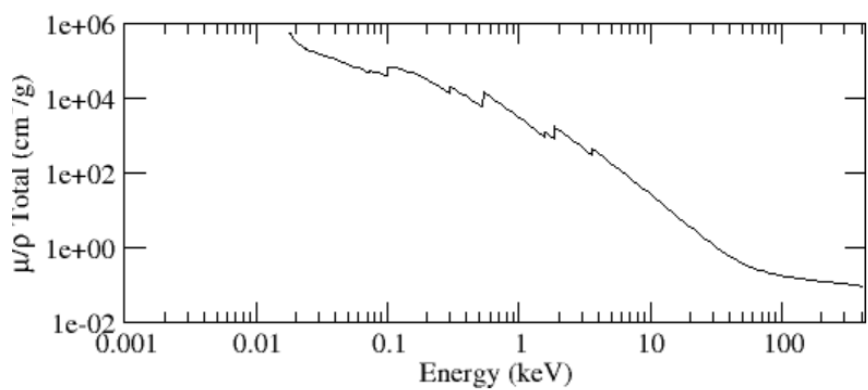